\documentclass[10pt,twocolumn,letterpaper]{article}
\usepackage{cvpr}
\usepackage{times}
\usepackage{epsfig}
\usepackage{graphicx}
\usepackage{amsmath}
\usepackage{amssymb}
\usepackage{amsmath}
\usepackage{amsthm}

\usepackage{algorithm}
\usepackage{algorithmicx}
\usepackage{algpseudocode}
\usepackage{booktabs}
\usepackage{float,comment,todonotes}
\usepackage{subcaption}
\usepackage{float}
\usepackage{dblfloatfix}

\usepackage[breaklinks=true,bookmarks=false]{hyperref}

\cvprfinalcopy 

\def\cvprPaperID{****} 

\begin{document}

\title{Structure-Preserving Physics-Informed Neural Network for the Korteweg–de Vries (KdV) Equation}
\author{%
   Victory C. Obieke\thanks{Oregon State University, obiekev@oregonstate.edu},\, \, 
   Emmanuel E. Oguadimma \thanks{Oregon State University, oguadime@oregonstate.edu} \, \,
}

\maketitle


\paragraph{Abstract.}
 Physics-Informed Neural Networks (PINNs) offer a flexible framework for solving nonlinear partial differential equations (PDEs), yet conventional implementations often fail to preserve key physical invariants during long-term integration. This paper introduces a structure-preserving PINN framework for the nonlinear Korteweg–de Vries (KdV) equation, a prototypical model for nonlinear and dispersive wave propagation.

The proposed method embeds conservation of mass and Hamiltonian energy directly into the loss function by designing a dynamic weighting mechanism for these invariants to improve long-term stability. We also develop a method for jointly learning these weights along with the network parameters, ensuring physically consistent and energy-stable evolution throughout both training and prediction.

To further contribute to the literature, we explore an understudied activation function—specifically, sinusoidal activations—which deviate from standard \texttt{tanh}-based PINNs~\cite{raissi2019pinn,wang2022modifiedpinn}. These sinusoidal units is seen to enhance spectral expressiveness and better capture the oscillatory and dispersive nature of KdV solitons.

Through representative case studies—including single-soliton propagation (shape-preserving translation), two-soliton interaction (elastic collision with phase shift), and cosine-pulse initialization (nonlinear dispersive breakup)—our model successfully reproduces hallmark behaviors of KdV dynamics while maintaining conserved quantities across time.\\
{\hskip 2em}

{\hskip 2em}

\section{Introduction}

Physics-Informed Neural Networks (PINNs) are a class of deep learning models that integrate physical laws directly into neural network architectures, enabling the solution of partial differential equations (PDEs) without the need for extensive labeled data. Introduced by Raissi \emph{et al.}~\cite{raissi2019pinn}, PINNs have since been applied successfully across a wide range of problems in scientific computing, including fluid dynamics, wave propagation, and nonlinear optics.

Classical numerical solvers such as the Finite Difference Method (FDM), Finite Element Method (FEM), and Fourier Pseudo-Spectral Method (PSM) have long been used to solve PDEs while maintaining the structural properties of their continuous formulations through careful discretization~\cite{TahaAblowitz1984JCP_KdV, FornbergWhitham1978, DeFrutosSanzSerna1992_JCP4th, CoxMatthews2002_ETD, KassamTrefethen2005_ETDRK4, Furihata1999_JCP_DVDM, FurihataMatsuo2011_Book, HoldenKarlsenRisebroTao2011_SplittingKdV}.
Despite their proven accuracy, these methods often face limitations when addressing high-dimensional, stiff, or long-time nonlinear dynamics, where discretization and stability constraints can become restrictive. PINNs overcome these challenges by embedding the governing equations directly into the learning process, combining the flexibility of data-driven models with the rigor of physics-based regularization~\cite{karniadakis2021physics, lin2022twostage, zhu2023modular}.

In this work, we propose a \emph{structure-preserving Physics-Informed Neural Network (SP-PINN)} for solving the nonlinear Korteweg–de Vries (KdV) equation—a canonical model of nonlinear and dispersive wave propagation. The KdV equation possesses an infinite hierarchy of conservation laws, with mass and Hamiltonian energy being the most fundamental. However, conventional PINNs do not explicitly enforce these invariants, leading to cumulative drift and degradation of physical fidelity during long-time integration.

To address this limitation, we incorporate conservation constraints for mass and energy directly into a unified single-stage loss function~\cite{BokilEtAl2026ICERM}, eliminating the need for multi-phase retraining as used in two-stage PINNs~\cite{lin2022twostage}. This formulation follows the philosophy of invariant-aware learning~\cite{wang2022modifiedpinn}, maintaining the Hamiltonian structure of the PDE while improving numerical stability and convergence efficiency.

A key innovation of our approach lies in coupling \textit{sinusoidal activation functions} with quasi-Newton optimization via the \textit{L-BFGS} algorithm. Sinusoidal activations enhance spectral expressiveness, naturally capturing the oscillatory, dispersive, and periodic characteristics of KdV solitons while mitigating spectral bias inherent in standard \texttt{tanh}-based networks. Although L-BFGS incurs a higher per-iteration computational cost, it converges rapidly to physically consistent minima with minimal hyperparameter tuning, making it well-suited for physics-informed optimization.

Together, these elements yield a compact, energy-consistent, and computationally efficient framework capable of accurately reproducing nonlinear dispersive dynamics, soliton interactions, and invariant preservation. The proposed SP-PINN thus offers a simple yet powerful paradigm for developing neural solvers that respect the intrinsic Hamiltonian structure of equations like the KdV and other nonlinear wave systems.

\section*{Acknowledgements}
The authors would like to thank the ICERM program
``Empowering a Diverse Computational Mathematics Research Community,''
and in particular the group on
``Compatible Discretizations for Nonlinear Optical Phenomena,''
where the idea of embedding the conservation of mass and Hamiltonian
energy directly into the loss function of the neural network originated.
We are grateful to the ICERM workshop group members Camille Carvalho,
Rafael Ceja Ayala, Maurice Fabien, Ruma Maity, and Claire Scheid for
their valuable discussions, and to our Ph.D. advisors, Vrushali Bokil
and Nathan Gibson, for their continuous guidance and support.


\section{Hamiltonian System}

A broad class of nonlinear dispersive equations, including the Korteweg–de Vries (KdV) equation, can be expressed in Hamiltonian form as
\begin{align*}
    u_t = \mathcal{D} \frac{\partial}{\partial u} \mathcal{H}, 
    \quad 
    \mathcal{H} = \int H\, dx,
\end{align*}
where $\mathcal{D}$ is a skew-adjoint differential operator, 
$\frac{\partial}{\partial u}$ denotes the variational derivative, and 
$\mathcal{H}$ represents the Hamiltonian functional with density $H$. 
By construction, the Hamiltonian quantity $\mathcal{H}$ is an invariant of the system, ensuring conservation of fundamental physical quantities such as mass and energy during temporal evolution.

Conventional Physics-Informed Neural Networks (PINNs) approximate PDE dynamics through residual minimization but do not inherently preserve these Hamiltonian invariants, often leading to drift in conserved quantities over long-time integration. 

\textbf{Objective:} 
To address this limitation, we develop a \emph{structure-preserving PINN} framework for the nonlinear KdV equation that explicitly embeds Hamiltonian conservation—specifically the invariance of mass and energy—within the learning objective, enabling physically consistent and long-term stable solutions.

  \section{KdV in 1D}
The Korteweg–de Vries (KdV) equation is a fundamental nonlinear partial differential equation that describes the evolution of long, shallow water waves with weakly nonlinear and dispersive effects. Originally derived to model shallow water waves, it has since been found to be applicable in a wide range of dispersive wave systems, including plasma physics, optical fibers, and even quantum field theory. We consider the KdV equation given by:
\begin{subequations}\label{KDV}
    \begin{align}
u_t + \eta u u_x &+ \mu^2 u_{xxx} = 0, \\
u(0,x) &= u_0(x) \\
u(t, a) &= 0, \, u(t, b) = 0, \quad a,b \in \mathbb{R}
\end{align}
\end{subequations}
where $u(t,x)$ represents the wave profile as a function of time $t \in [0, T]$ and spatial position $x \in [a,b]$, $\eta \in \mathbb{R}$ is the nonlinearity parameter, $\mu$ is the dispersion coefficient. This equation exhibits soliton solutions, which are localized waves that maintain their shape while propagating at a constant velocity. This remarkable property has made the KdV equation one of the most studied nonlinear wave equations.

\subsection{Physical Properties of KDV}
We consider the the scaled KDV equation \eqref{KDV} posed on $\mathbb{R}$, with $u \to 0$ as $|x| \to \infty$. Under these boundary conditions, the KDV equation is Hamiltonian and admits an infinite hierarchy of conserved quantities \cite{miura1968korteweg}. In particular, the "mass" and the Hamiltonian ("energy") are conserved. 

\subsubsection*{Mass}
We define the mass $\mathcal{M}(t)$ by 
\begin{align}\label{mass}
    \mathcal{M}(t) &= \int_{\mathbb{R}} u(x,t) \, dx
\end{align}
Then we have 
\begin{align*}
    \frac{ d\mathcal{M}(t)}{dt} &= \frac{d}{dt}\int_{\mathbb{R}} u(x,t) \, dx= \int_{\mathbb{R}} \partial_t u(x,t) \, dx \\
    &= \int_{\mathbb{R}} -\partial_x \left(\frac{\eta}{2} u^2 + \mu^2 u_{xx}\right) \, dx \\
    &= \left[\frac{\eta}{2} u^2 + \mu^2 u_{xx}\right]_{x = -\infty}^{x = \infty}\\
    &= 0
\end{align*}
Hence, $\mathcal{M}(t) = \text{constant}$ for all $t$, implying that the mass is conserved for the KDV.  

\subsubsection*{Energy/Hamiltonian}
We define the energy (Hamiltonian) $\mathcal{E}(t)$ of the solution as 
\begin{align}\label{energy}
    \mathcal{E} = \int_{\mathbb{R}} \left(\frac{\mu^2}{2}u_x^2 - \frac{\eta}{6}u^3\right) dx
\end{align}
Set 
\[
  f = \frac{\eta}{2}u^2 + \mu^2 u_{xx}  
\]
Then, we have 
\[
    f_x = \eta u u_x + \mu^2 u_{xxx}.
\]
From the energy functional
\[
 \mathcal{E} = \int \left( \frac{\mu^2}{2}u_x^2 - \frac{\eta}{6}u^3 \right)\,dx,
\]
and the KdV equation \(u_t = -f_x\), we have
\[
\frac{d \mathcal{E}}{dt}
= \int \left( -\mu^2 u_x f_{xx} + \frac{\eta}{2}u^2 f_x \right)\,dx.
\]
Integrating the first term by parts once gives
\[
-\mu^2 \int u_x f_{xx}\,dx
= -\mu^2 [u_x f_x]_{-\infty}^{\infty}
+ \mu^2 \int u_{xx} f_x\,dx.
\]
Therefore,
\begin{align*}
    \frac{d \mathcal{E}}{dt} &= -\mu^2 [u_x f_x]_{-\infty}^{\infty}
    + \int \left( \mu^2 u_{xx} + \frac{\eta}{2}u^2 \right) f_x\,dx \\
    &= -\mu^2 [u_x f_x]_{-\infty}^{\infty}
    + \int f\,f_x\,dx \\
    &=  -\mu^2 [u_x f_x]_{-\infty}^{\infty} + \frac{1}{2}\int \partial_x(f^2)\,dx\\
    &= -\mu^2 [u_x f_x]_{-\infty}^{\infty} + \left[ \frac{1}{2} f^2 \right]_{-\infty}^{\infty} \\
    &= 0
\end{align*}
Hence, $\mathcal{E}(t) = \text{constant}$ for all $t$, showing that the KdV energy (Hamiltonian) is conserved.

\section{Methodology}

We outline the methods employed to achieve a stable and physically consistent solution of the KdV equation using physics-informed neural networks (PINNs). The proposed framework preserves the physical invariants of the model while ensuring numerical stability throughout training and inference.

\begin{figure}[H]
    \centering
    \includegraphics[width=1\linewidth]{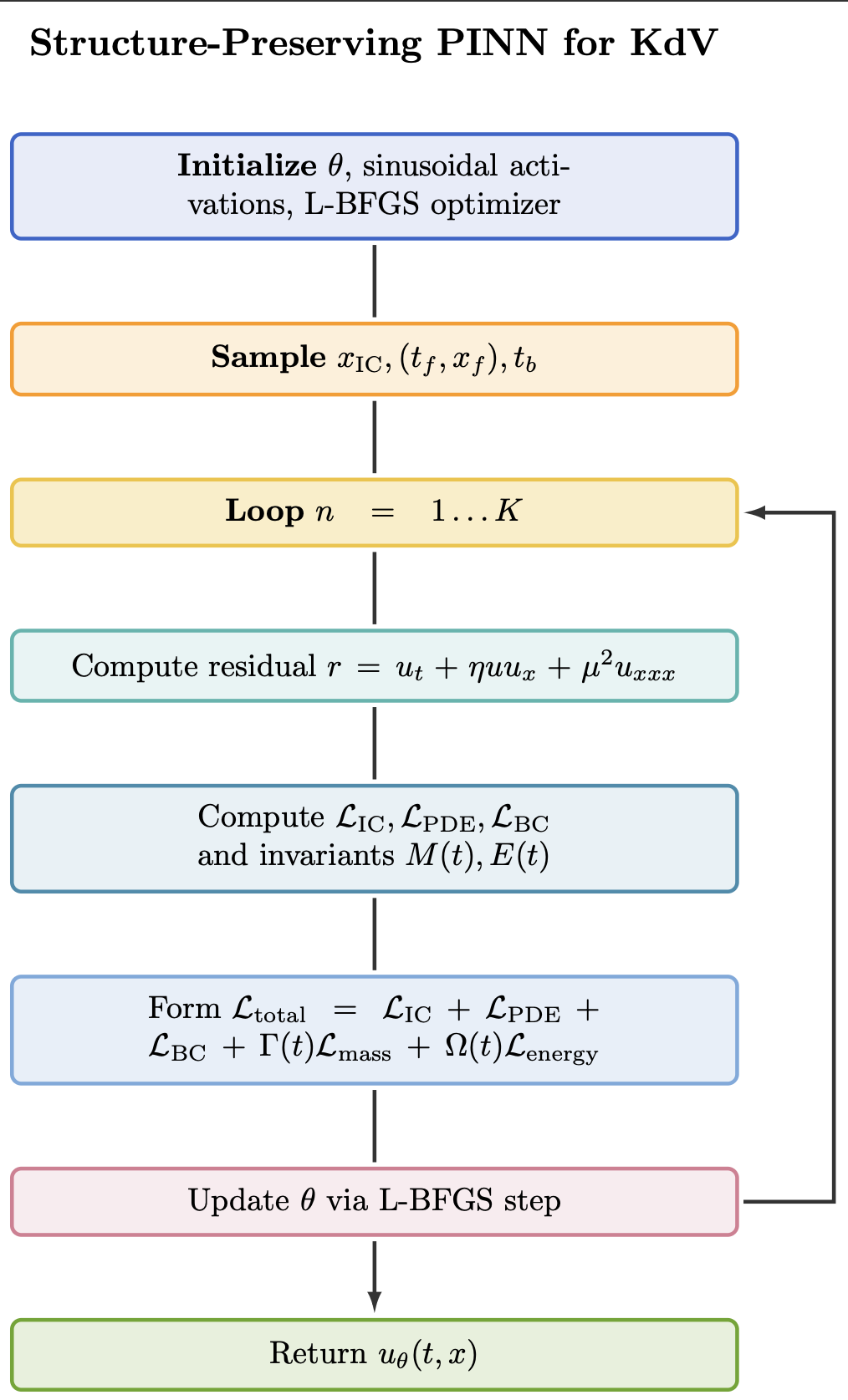}
    \caption{Structure-Preserving PINN Framework for KdV}
    \label{fig:kdv_flow}
\end{figure}

\subsection{Neural Network Architecture}

As in ~\ref{fig:kdv_flow}, we model the solution $u(t,x)$ using a fully-connected feedforward neural network (FNN), denoted by $u_\theta(t,x)$ and parameterized by the trainable weights $\theta$. The network takes the two-dimensional input $(t, x)$ and outputs the scalar field $u(t,x)$. A sinusoidal activation function $\sin(x)$ is employed in all hidden layers to capture the oscillatory and dispersive characteristics of soliton dynamics. The network architecture is defined as follows:
\begin{itemize}
    \item \textbf{Input layer:} 2 neurons corresponding to $(t, x)$,
    \item \textbf{Hidden layers:} $L$ fully-connected layers of width $W$ with sinusoidal activation,
    \item \textbf{Output layer:} 1 neuron producing the scalar output $u_\theta(t,x)$.
\end{itemize}

\subsection{Loss Function Components}

The total loss $\mathcal{L}_{\text{total}}(\theta)$ comprises multiple components that jointly enforce the governing dynamics, boundary conditions, and conservation properties of the KdV system:

\begin{enumerate}
    \item \textbf{Initial Condition Loss:}
    \[
    \mathcal{L}_{\text{IC}} = \frac{1}{N_{\text{IC}}} \sum_{i=1}^{N_{\text{IC}}} 
    \left| u_\theta(0, x_i) - u_0(x_i) \right|^2,
    \]
    where $u_0(x)$ represents the known analytical initial condition from the soliton profile.

    \item \textbf{PDE Residual Loss:}
    The KdV dynamics are enforced via automatic differentiation \cite{raissi2019pinn} at collocation points $(t_i, x_i)$:
    \[
    \mathcal{L}_{\text{PDE}} = \frac{1}{N_f} \sum_{i=1}^{N_f}
    \left| \partial_t u_\theta + 6 u_\theta \partial_x u_\theta + \partial_{xxx} u_\theta \right|^2.
    \]

    \item \textbf{Boundary Condition Loss:}
    Homogeneous Dirichlet boundary conditions are imposed on the spatial boundaries:
    \[
    \mathcal{L}_{\text{BC}} = \frac{1}{N_b} \sum_{i=1}^{N_b}
    \left( |u_\theta(t_i, x_{\min})|^2 + |u_\theta(t_i, x_{\max})|^2 \right).
    \]

    \item \textbf{Conservation Laws:}
    To promote physical fidelity, we include soft constraints on the conservation of mass and energy:
    \begin{align*}
        \mathcal{L}_{\text{mass}} &= \frac{1}{N_t} \sum_{j=1}^{N_t}
        \left| \mathcal{M}(t_j) - \mathcal{M}(0) \right|^2, \\
        \mathcal{L}_{\text{energy}} &= \frac{1}{N_t} \sum_{j=1}^{N_t}
        \left| \mathcal{E}(t_j) - \mathcal{E}(0) \right|^2,
    \end{align*}
    where $\mathcal{M}$ and $\mathcal{E}$ denote the total mass and energy defined in
    Eqs.~\eqref{mass} and~\eqref{energy}, respectively, with parameters
    $\mu = 1$ and $\eta = 6$.
\end{enumerate}

\subsection{Dynamic Weighting of Invariant Loss Terms}

The total composite loss is defined as
\begin{equation}
\mathcal{L}_{\text{total}} =
\mathcal{L}_{\text{IC}} +
\mathcal{L}_{\text{PDE}} +
\mathcal{L}_{\text{BC}} +
\Gamma(t)\,\mathcal{L}_{\text{mass}} +
\Omega(t)\,\mathcal{L}_{\text{energy}},
\end{equation}
where $\Gamma(t)$ and $\Omega(t)$ are \textit{dynamically adjusted coefficients} that control the relative contribution of the mass and energy conservation terms during training.

The coefficient $\Gamma(t)$ acts as a \textit{mass-balancing weight}, increasing when deviations from mass conservation grow and decreasing once the invariant stabilizes. 
Similarly, $\Omega(t)$ serves as an \textit{energy-balancing weight}, adapting to ensure that the Hamiltonian energy constraint contributes proportionally to the total loss. 
Together, these coefficients enforce \textit{structure preservation}, ensuring that the neural solution satisfies both the governing PDE and its underlying physical invariants.

Dynamic weighting strategy is based on \textit{gradient normalization} defined as
\begin{align*}
\Gamma(t) =
\frac{\|\nabla_\theta \mathcal{L}_{\text{PDE}}\|_2}
{\|\nabla_\theta \mathcal{L}_{\text{mass}}\|_2 + \varepsilon},
\qquad
\Omega(t) =
\frac{\|\nabla_\theta \mathcal{L}_{\text{PDE}}\|_2}
{\|\nabla_\theta \mathcal{L}_{\text{energy}}\|_2 + \varepsilon},
\end{align*}
where $\nabla_\theta$ denotes the gradient with respect to the network parameters, and $\varepsilon$ is a small stabilization constant.
This formulation ensures that each loss component contributes comparably to the overall gradient magnitude, preventing either the PDE residual or invariant terms from dominating the optimization process.

\subsection{Training Procedure}

The parameters $\theta$ are optimized using the L-BFGS algorithm, a quasi-Newton method well-suited for stiff loss landscapes and high-dimensional PDE constraints. All derivatives and residual terms are computed using automatic differentiation in PyTorch. The optimizer operates within a closure-style loop, allowing precise gradient evaluation at each iteration. Convergence is typically achieved within a few thousand iterations.

\vskip0.5ex
\noindent
\textbf{Implementation details.} 
Unless otherwise specified, the network consists of $L = 4$ hidden layers with width $W = 40$, trained using the L-BFGS optimizer (learning rate $1$, \texttt{max\_iter} $= 3000$). The number of collocation and boundary sampling points are set to $N_f = 8192$ and $N_b = 128$, respectively. These hyperparameters were chosen to balance training stability, accuracy, and computational cost.

\vskip0.5ex
\noindent
\textbf{Computational setup.} 
All experiments were performed in Python~3.12 using PyTorch~2.3. The simulations ran on a single NVIDIA RTX~A5000 GPU with 24\,GB VRAM and 64\,GB system memory. Each full training time varied depending on the resolution of collocation points and the size of the computational domain. Visualization and post-processing were conducted using \texttt{Matplotlib} and \texttt{NumPy}.
\begin{table}[H]
\centering
\caption{Training configuration and computational cost for each KdV test case (RTX~A5000 GPU, 24~GB VRAM).}
\begin{tabular}{lccccc}
\toprule
Case & Layers & Width & Iterations & Time (min) \\
\midrule
One-soliton  & 4 & 40 & 3000 & 8\\
Two-soliton  & 7 & 40 & 3000 & 22 \\
Cosine pulse & 4 & 40 & 3000 & 6 \\
\bottomrule
\end{tabular}
\end{table}
\footnotesize{All cases used the L-BFGS optimizer with an initial learning rate of~1.0.}

\subsection{Post-Training Evaluation}

After training, the neural network prediction $u_\theta(t,x)$ is evaluated over a fine spatio-temporal mesh and compared against the analytical two-soliton solution. The following diagnostics are used to assess model performance:
\begin{itemize}
    \item Comparison of predicted versus analytical solution profiles at selected time instances,
    \item Space–time surface visualizations of $u(t,x)$,
    \item Two-dimensional contour maps of the absolute prediction error,
    \item Temporal evolution of conserved quantities (mass and energy).
\end{itemize}
All figures are generated using \texttt{Matplotlib} and stored for reproducibility.

\subsection{Algorithm}
\begin{algorithm}[H]
\caption{Structure-Preserving PINN for KdV}
\begin{algorithmic}[1]
\Require Domains $[0,T]$, $[x_{\min},x_{\max}]$; network $u_\theta$; max iterations $K$
\Ensure Trained $u_\theta$ approximating KdV
\State Initialize $\theta$ (sinusoidal activations), L-BFGS optimizer
\State Sample $x_{\mathrm{IC}}$, interior $(t_f,x_f)$, boundary $t_b$
\For{$n=1$ to $K$}
  \State Compute residual $r = u_t + 6uu_x + u_{xxx}$ at $(t_f,x_f)$
  \State Compute $\mathcal{L}_{\mathrm{IC}}, \mathcal{L}_{\mathrm{PDE}}, \mathcal{L}_{\mathrm{BC}}$
  \State Estimate $M(t),E(t)$ and compute $\mathcal{L}_{\mathrm{mass}}, \mathcal{L}_{\mathrm{energy}}$
  \State $\mathcal{L}_{\mathrm{total}} \leftarrow \mathcal{L}_{\mathrm{IC}}+\mathcal{L}_{\mathrm{PDE}}+\mathcal{L}_{\mathrm{BC}}+\Gamma(t)\mathcal{L}_{\mathrm{mass}}+\Omega(t) \mathcal{L}_{\mathrm{energy}}$
  \State Update $\theta$ via optimizer
\EndFor
\State \Return $u_\theta$
\end{algorithmic}
\end{algorithm}

\section{Results}
In this section, we evaluate the performance of our physics-informed neural network (PINN) by solving the Korteweg--de Vries (KdV) equation on a bounded domain. We first consider exact initial and boundary conditions derived from a known soliton solution, and finally, we consider exact initial and boundary conditions defined in \cite{wang2022modifiedpinn}.  

\

\noindent 
\textit{One-Soliton Profile.}
\
\vskip1ex
\noindent
We first consider the nondimensional Korteweg–de Vries (KdV) equation~\eqref{KDV} with parameters $\mu = 1$ and $\eta = 6$, defined on the domain $(t, x) \in [0, 3] \times [-20, 20]$. Following the setup in~\cite{dag2008kdv}, the initial condition corresponds to the exact one-soliton profile:
\[
    u(0, x) = \frac{c}{2}\,\text{sech}^2\!\left( \frac{\sqrt{c}}{2}(x - x_0) \right),
\]
where the soliton has speed $c = 1$ and initial center $x_0 = 0$. The corresponding analytical solution of the KdV equation is given by
\begin{equation}
    u(t, x) = \frac{c}{2}\,\operatorname{sech}^2\!\left( \frac{\sqrt{c}}{2}\,(x - c t - x_0) \right),
    \label{eq:kdv_soliton}
\end{equation}
where \( c > 0 \) denotes the wave speed.  
This test case provides a fundamental benchmark for evaluating the model’s ability to reproduce solitary-wave propagation with correct amplitude, shape, and translation speed, key signatures of the integrable KdV dynamics.

\
\ 

\noindent 
\textit{Two Soliton Profile.}
\ 
\vskip1ex 
\noindent 
We consider the nondimensional Korteweg–de Vries (KdV) equation~\eqref{KDV} with parameters $\mu = 1$ and $\eta = 6$, defined on the domain $(t, x) \in [0, 10\pi] \times [-40, 40]$. Following the setup in~\cite{dag2008kdv}, the initial condition is chosen as a nonlinear superposition of two solitary waves (solitons) with distinct amplitudes and propagation speeds:
\begin{align*}
    u(0, x) &= \frac{c_1}{2} \, \text{sech}^2\!\left( \frac{\sqrt{c_1}}{2}(x - x_1)\right) \\
             &\quad + \frac{c_2}{2} \, \text{sech}^2\!\left( \frac{\sqrt{c_2}}{2}(x - x_2)\right),
\end{align*}

where $c_1 = 1$, $c_2 = 0.3$, $x_1 = -5$, and $x_2 = 5$.  
This configuration generates two solitons of different heights and velocities that propagate toward each other, undergo an elastic collision, and then re-emerge without change in shape or speed—exhibiting the hallmark dispersive and nonlinear characteristics of the KdV dynamics. This two-soliton interaction serves as a canonical benchmark for assessing the ability of numerical or learning-based models to capture nonlinear wave propagation and soliton interactions~\cite{dag2008kdv}.

\

\ 

\noindent 
\textit{Cosine Profile.}
\vskip1ex
\noindent
To further demonstrate the robustness of our model, we consider a setup similar to that in~\cite{wang2022modifiedpinn,dag2008kdv}, which examines the breakup of an initial smooth pulse into a train of solitons. The initial condition is given by
\[
    u(0, x) = \cos(\pi x),
\]
defined on the domain $(t, x) \in [0, 1] \times [-1, 1]$. We simulate the nondimensional KdV equation~\eqref{KDV} with parameters $\eta = 1$ and $\mu = 0.05$. The resulting evolution shows that our model accurately reproduces the pulse breakup dynamics reported in the literature, confirming its capability to capture complex dispersive behaviors and nonlinear wave interactions inherent to the KdV system.

\subsection{Ablation Study: Impact of Conservation Laws on PINN Performance}
In this section, we investigate the impact of embedding the conservation laws into the loss function. 
\begin{table}[H]
\centering
\resizebox{0.45\textwidth}{!}{%
\begin{tabular}{cccccccc}
\toprule
 & \multicolumn{3}{c}{\textbf{SP-PINN}} &  & \multicolumn{3}{c}{\textbf{Vanilla PINN}} \\
\cmidrule(lr){2-4} \cmidrule(lr){6-8}
\textbf{Time} & \textbf{Mass} & \textbf{Energy} & \textbf{Error} &  & \textbf{Mass} & \textbf{Energy} & \textbf{Error} \\
\midrule
0.00 & 1.9997 & $-0.1998$ & $3.36\times10^{-4}$ &  & 2.0000 & $-0.2000$ & $1.35\times10^{-4}$ \\
0.05 & 2.0002 & $-0.1999$ & $3.57\times10^{-4}$ &  & 2.0000 & $-0.2001$ & $1.47\times10^{-4}$ \\
0.10 & 2.0006 & $-0.2000$ & $4.01\times10^{-4}$ &  & 2.0000 & $-0.2001$ & $1.67\times10^{-4}$ \\
0.60 & 1.9995 & $-0.1999$ & $5.79\times10^{-4}$ &  & 1.9998 & $-0.2002$ & $2.55\times10^{-4}$ \\
0.80 & 1.9987 & $-0.1997$ & $5.77\times10^{-4}$ &  & 1.9995 & $-0.2001$ & $2.77\times10^{-4}$ \\
1.00 & 1.9986 & $-0.1997$ & $6.28\times10^{-4}$ &  & 1.9990 & $-0.2001$ & $3.18\times10^{-4}$ \\
1.40 & 2.0004 & $-0.1998$ & $7.83\times10^{-4}$ &  & 1.9975 & $-0.2002$ & $5.85\times10^{-4}$ \\
1.80 & 2.0011 & $-0.1999$ & $7.77\times10^{-4}$ &  & 1.9951 & $-0.2003$ & $1.07\times10^{-3}$ \\
2.00 & 1.9984 & $-0.1995$ & $7.58\times10^{-4}$ &  & 1.9931 & $-0.2003$ & $1.30\times10^{-3}$ \\
2.20 & 1.9992 & $-0.1997$ & $8.01\times10^{-4}$ &  & 1.9904 & $-0.2004$ & $2.95\times10^{-3}$ \\
2.40 & 1.9998 & $-0.1998$ & $8.42\times10^{-4}$ &  & 1.9880 & $-0.2005$ & $4.76\times10^{-3}$ \\
2.60 & 2.0000 & $-0.1998$ & $8.73\times10^{-4}$ &  & 1.9852 & $-0.2006$ & $6.89\times10^{-3}$ \\
2.80 & 2.0003 & $-0.1999$ & $9.10\times10^{-4}$ &  & 1.9835 & $-0.2007$ & $8.95\times10^{-3}$ \\
3.00 & 2.0001 & $-0.1999$ & $9.45\times10^{-4}$ &  & 1.9819 & $-0.2008$ & $1.12\times10^{-2}$ \\
\bottomrule
\end{tabular}
}
\caption{Comparison of temporal conservation performance between the  SP-PINN and Vanilla PINN for the KdV equation (learning rate = 0.1, sine activation). The SP-PINN maintains invariant errors within $10^{-4}$ up to $t=3$, while the Vanilla PINN drifts toward $10^{-2}$, highlighting the superior long-term stability of the proposed approach.}
\label{tab:l1}
\end{table}

\begin{table}[H]
\centering
\resizebox{0.45\textwidth}{!}{%
\begin{tabular}{cccccccc}
\toprule
 & \multicolumn{3}{c}{\textbf{SP-PINN}} &  & \multicolumn{3}{c}{\textbf{Vanilla PINN}} \\
\cmidrule(lr){2-4} \cmidrule(lr){6-8}
\textbf{Time} & \textbf{Mass} & \textbf{Energy} & \textbf{Error} &  & \textbf{Mass} & \textbf{Energy} & \textbf{Error} \\
\midrule
0.00 & 1.9997 & $-0.2000$ & $1.40\times10^{-4}$ &  & 1.9992 & $-0.1999$ & $2.79\times10^{-4}$ \\
0.05 & 2.0000 & $-0.2000$ & $1.29\times10^{-4}$ &  & 1.9989 & $-0.1999$ & $2.68\times10^{-4}$ \\
0.10 & 2.0001 & $-0.2000$ & $1.42\times10^{-4}$ &  & 1.9987 & $-0.1998$ & $2.79\times10^{-4}$ \\
0.60 & 1.9999 & $-0.2000$ & $1.86\times10^{-4}$ &  & 1.9968 & $-0.1997$ & $5.29\times10^{-4}$ \\
0.80 & 1.9995 & $-0.2000$ & $1.53\times10^{-4}$ &  & 1.9959 & $-0.1998$ & $6.43\times10^{-4}$ \\
1.00 & 1.9992 & $-0.2000$ & $1.50\times10^{-4}$ &  & 1.9949 & $-0.1999$ & $7.89\times10^{-4}$ \\
1.40 & 1.9994 & $-0.1999$ & $2.29\times10^{-4}$ &  & 1.9925 & $-0.1999$ & $1.17\times10^{-3}$ \\
1.80 & 2.0000 & $-0.2000$ & $2.56\times10^{-4}$ &  & 1.9891 & $-0.1995$ & $1.66\times10^{-3}$ \\
2.00 & 1.9998 & $-0.2000$ & $2.69\times10^{-4}$ &  & 1.9861 & $-0.1992$ & $1.93\times10^{-3}$ \\
2.20 & 2.0000 & $-0.2000$ & $2.95\times10^{-4}$ &  & 1.9839 & $-0.1991$ & $4.28\times10^{-3}$ \\
2.40 & 2.0002 & $-0.2000$ & $3.22\times10^{-4}$ &  & 1.9810 & $-0.1990$ & $7.63\times10^{-3}$ \\
2.60 & 2.0003 & $-0.2000$ & $3.48\times10^{-4}$ &  & 1.9778 & $-0.1988$ & $1.03\times10^{-2}$ \\
2.80 & 2.0002 & $-0.2000$ & $3.79\times10^{-4}$ &  & 1.9749 & $-0.1986$ & $1.07\times10^{-2}$ \\
3.00 & 2.0001 & $-0.2000$ & $4.05\times10^{-4}$ &  & 1.9725 & $-0.1984$ & $1.11\times10^{-2}$ \\
\bottomrule
\end{tabular}
}
\caption{Comparison of temporal conservation performance between the  SP-PINN and Vanilla PINN for the KdV equation (learning rate =1, sine activation). The SP-PINN maintains invariant errors within $10^{-4}$ up to $t=3$, while the regular PINN exhibits error growth into the $10^{-2}$ range, confirming its weaker long-term stability.}
\label{tab:l1i}
\end{table}

Tables~\ref{tab:l1}--\ref{tab:l1i} compare the errors produced by our approach and the vanilla PINN (without conservation laws) for learning rates $\gamma = 0.1$ and $\gamma = 1$, respectively. We observe a clear deterioration in the accuracy of the vanilla PINN as time advances, reflecting its inability to maintain physical invariants over long horizons. In contrast, our structure-preserving PINN maintains stable and accurate predictions throughout the same period. Notably, our model converges even for a large learning rate ($\gamma = 1$), whereas the baseline diverges. This suggests that the sinusoidal activation function provides superior numerical stability and spectral representation compared to the widely used \texttt{tanh} activation \cite{raissi2019pinn,wang2022modifiedpinn}.

\ 
\

\subsection{Numerical Simulations}

\begin{figure}[H]
    \centering
    \includegraphics[width=1\linewidth]{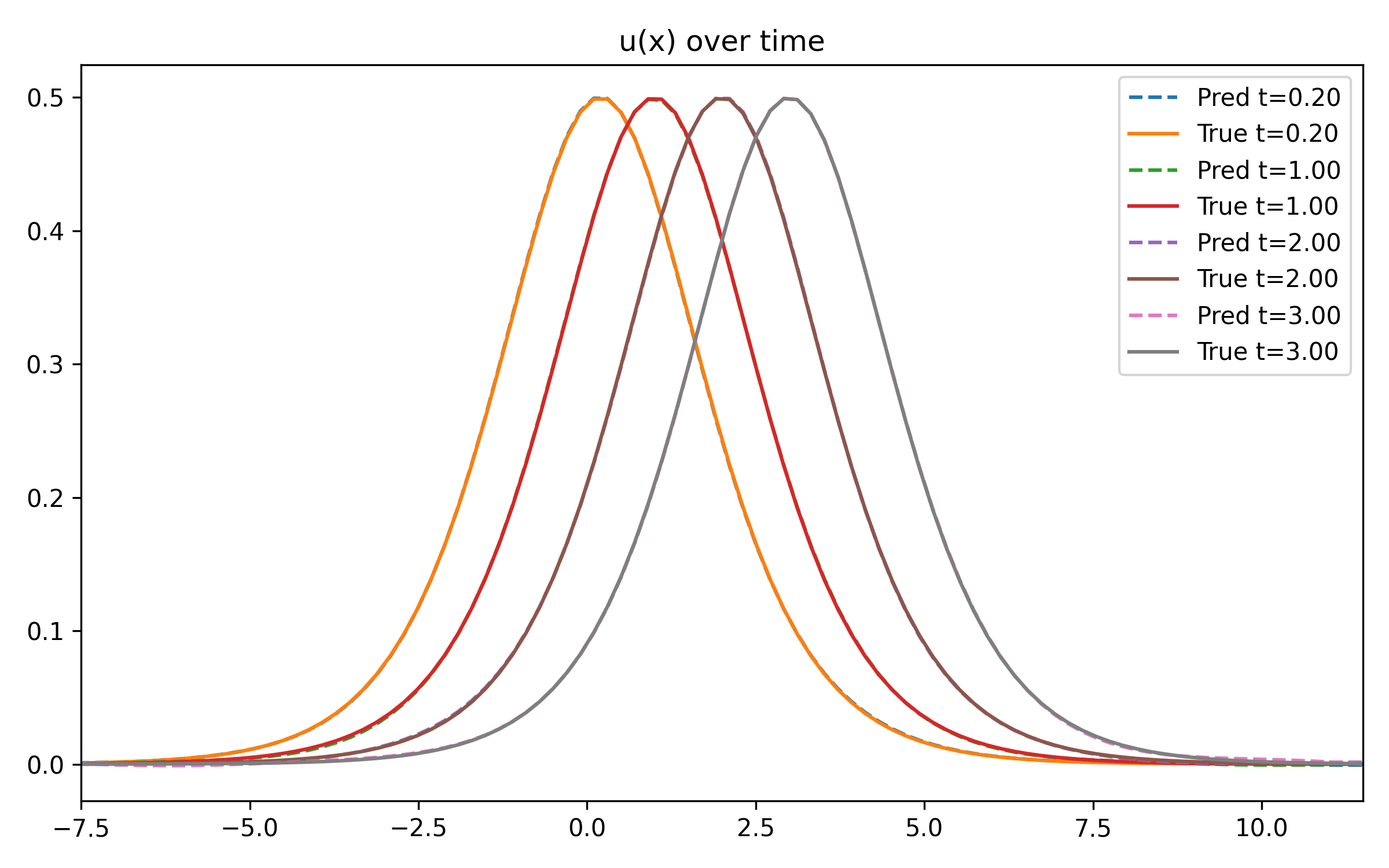}
    \caption{Comparison of predicted vs. exact soliton profiles at various time snapshots. The dashed lines denote PINN predictions; solid lines denote the true solution.}
    \label{fig:kdv_profiles1}
\end{figure}

\begin{figure}[H]
    \centering
    \includegraphics[width=1\linewidth]{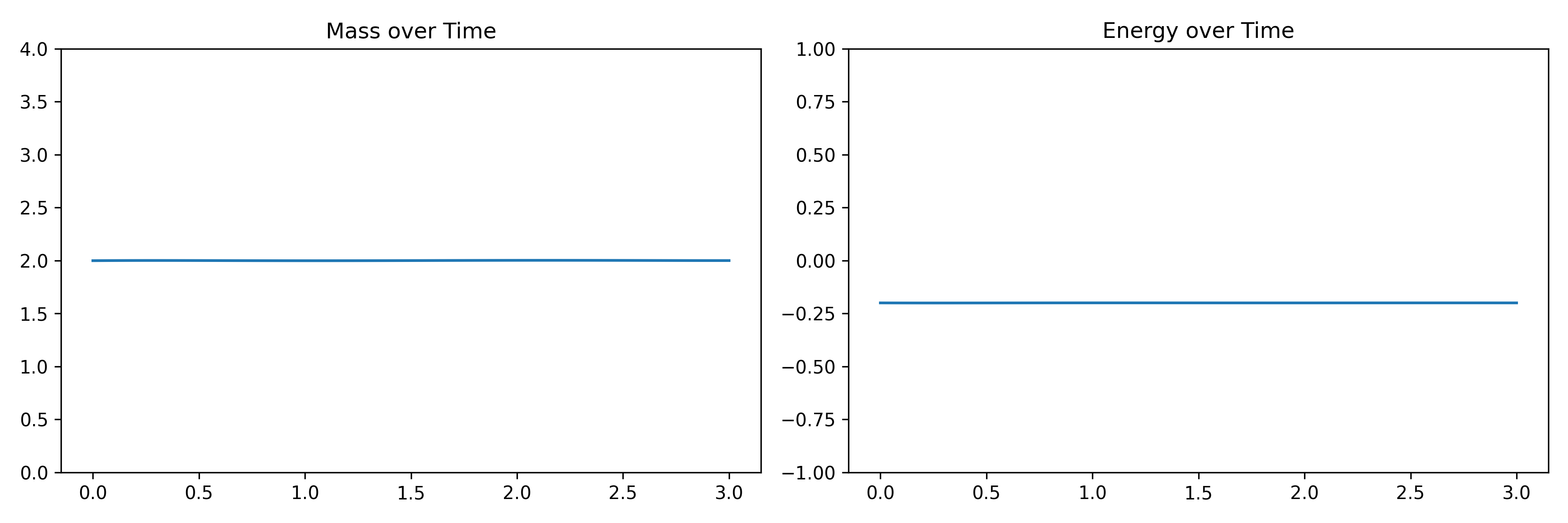}
    \caption{Mass (left) and energy (right) conservation over time for the PINN solution for one soliton profile.}
    \label{fig:mass_energy1}
\end{figure}

\begin{figure*}[t]
    \centering
    \includegraphics[width=\textwidth]{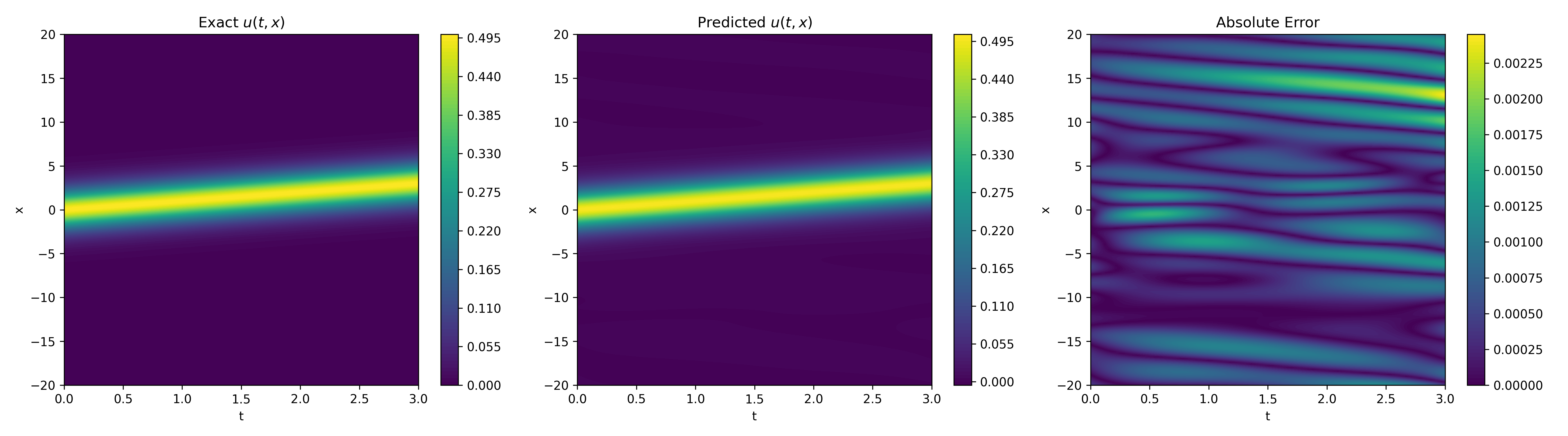}
    \caption{Contour plots of the exact solution $u(t,x)$ (left), PINN-predicted solution (middle), and absolute error (right).}
    \label{fig:contours1}
\end{figure*}

\vspace{1ex}
\noindent\textbf{\textit{One-Soliton Profile.}} 
The one-soliton solution of the Korteweg–de Vries (KdV) equation represents a solitary wave that maintains its amplitude and shape as it travels at a constant speed. This equilibrium between nonlinearity and dispersion produces a stable, self-reinforcing waveform. The contour and profile plots in Fig.~\ref{fig:kdv_profiles1}, ~\ref{fig:contours1}confirm that the trained PINN successfully reproduces this behavior, closely matching the analytical solution with minimal absolute error across time which is smaller when compared with \cite{wang2022modifiedpinn} with physically conserved quantities in ~\ref{fig:mass_energy1}

\begin{figure}[H]
\centering
\fbox{%
\begin{minipage}{0.9\linewidth}
\centering
\begin{tabular}{cc}
\includegraphics[width=0.45\linewidth]{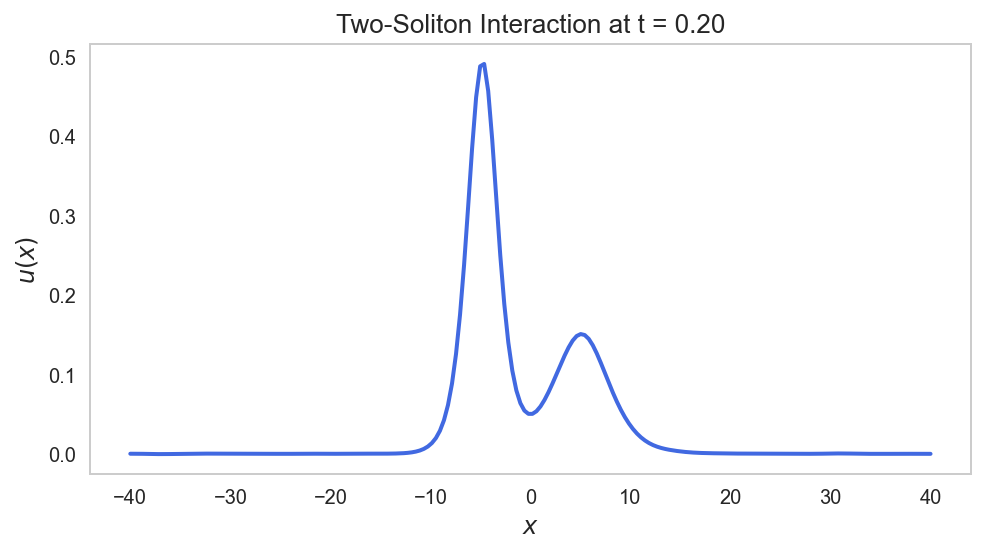} &
\includegraphics[width=0.45\linewidth]{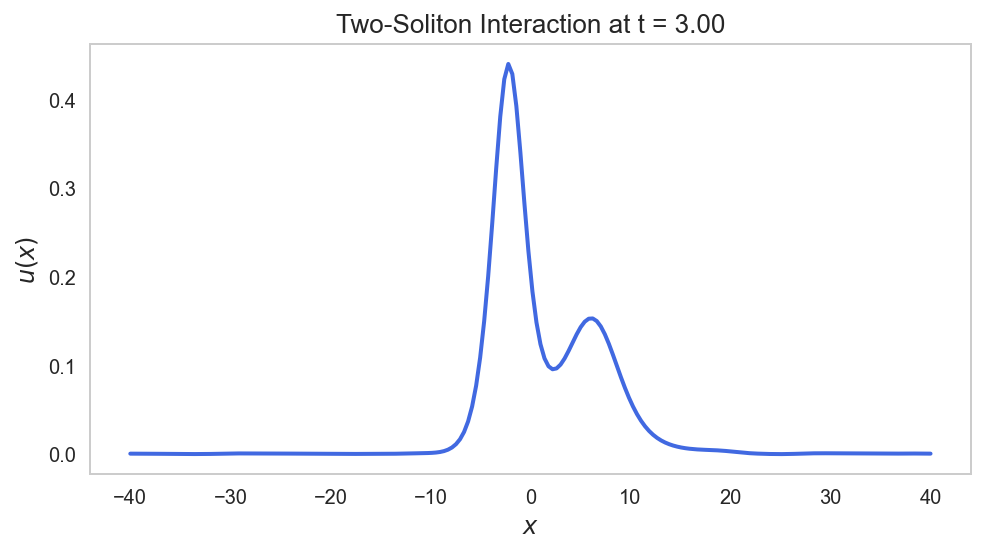} \\[6pt]
\includegraphics[width=0.45\linewidth]{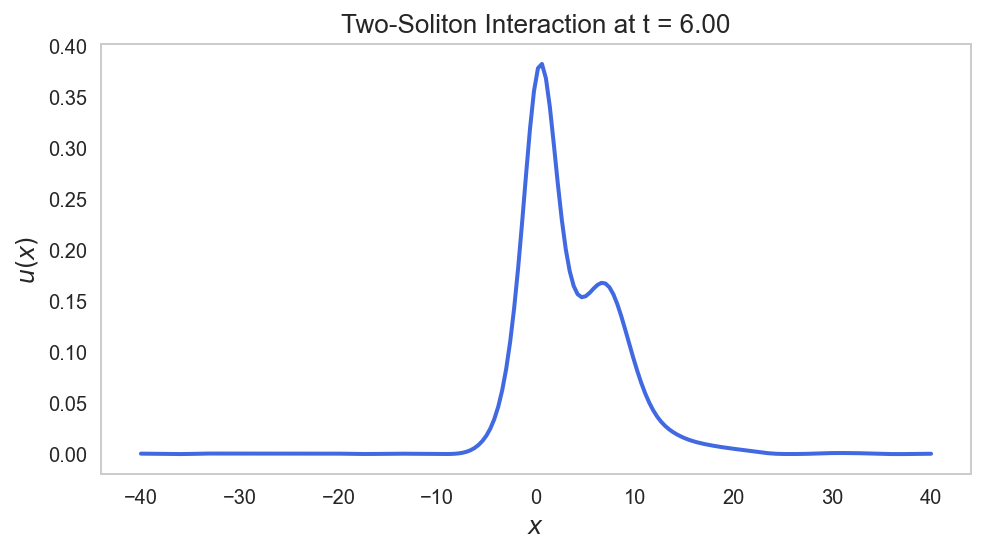} &
\includegraphics[width=0.45\linewidth]{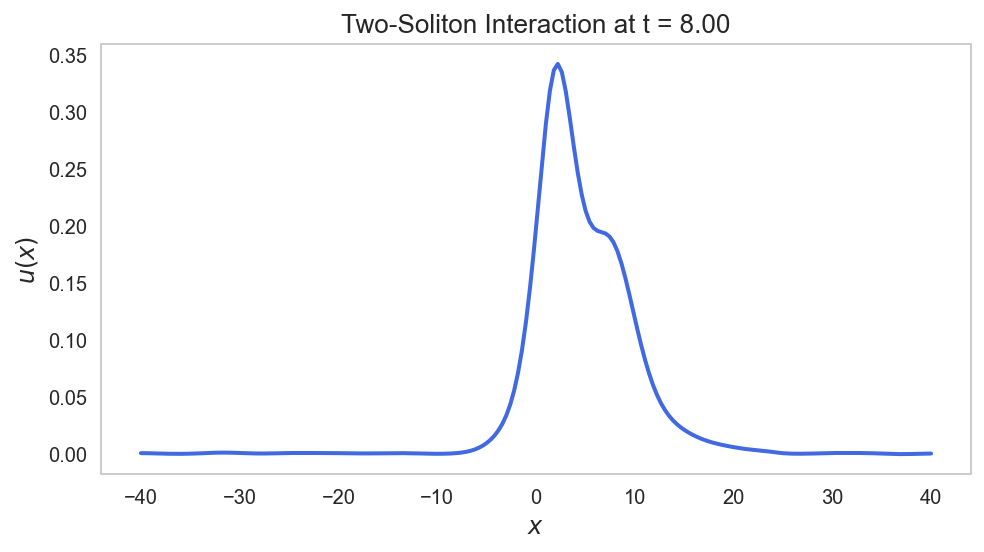} \\[6pt]
\includegraphics[width=0.45\linewidth]{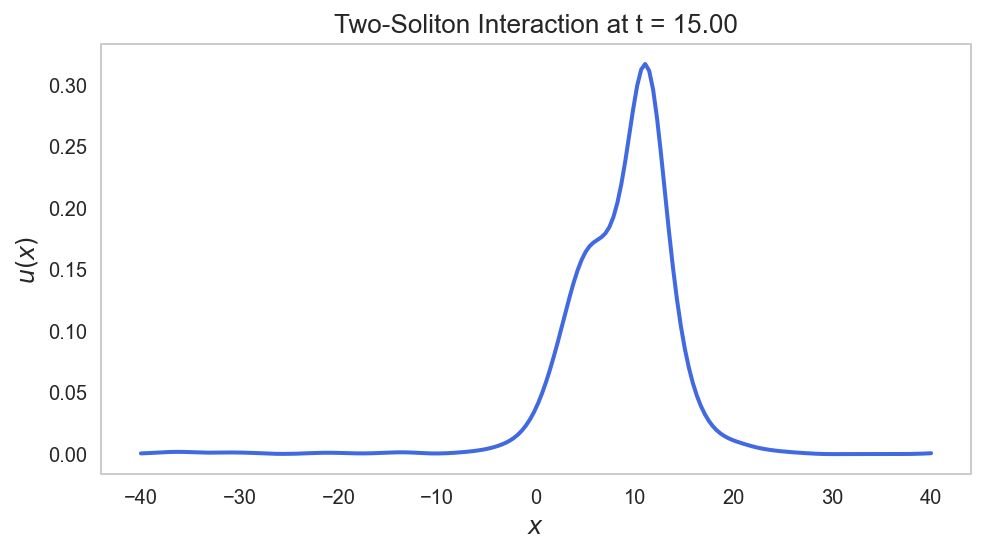} &
\includegraphics[width=0.45\linewidth]{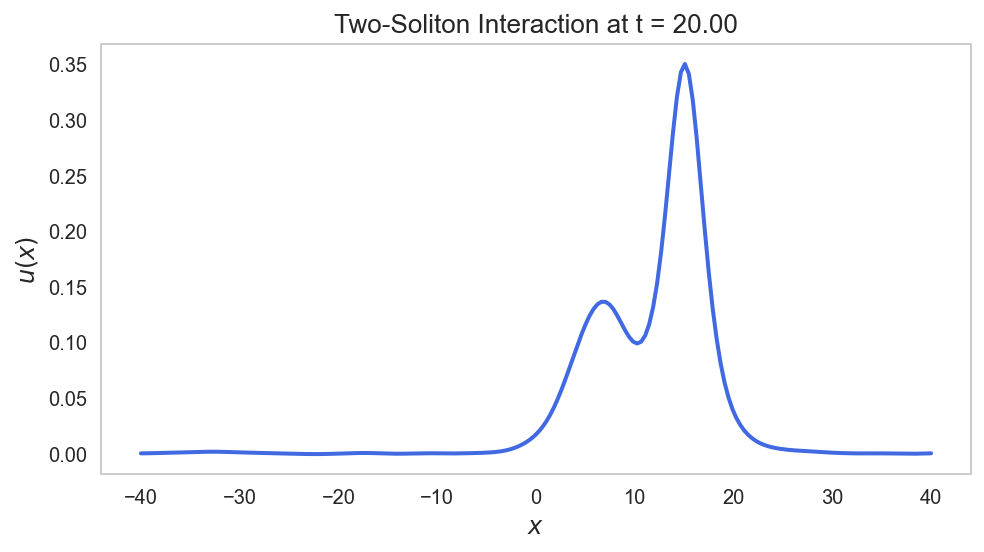} \\[6pt]
\includegraphics[width=0.45\linewidth]{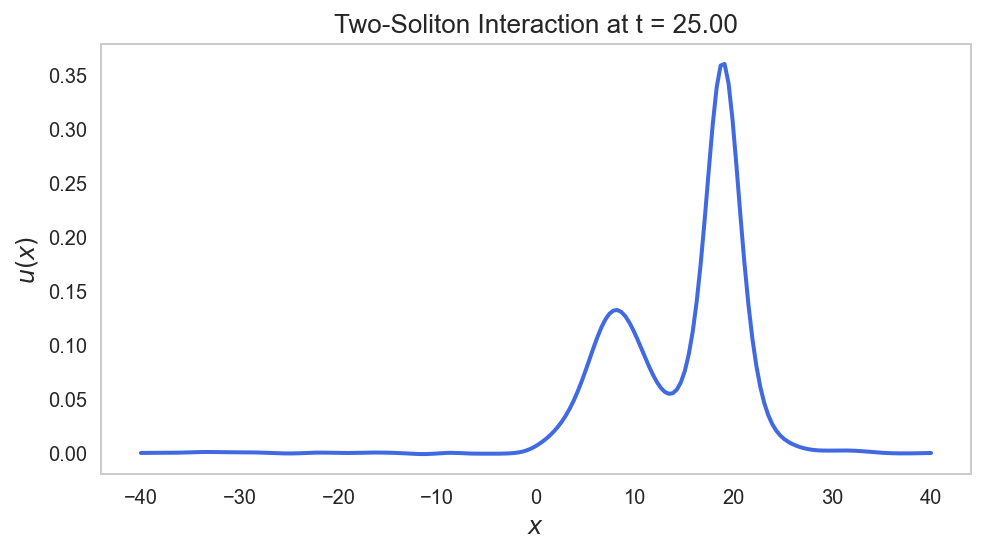} &
\includegraphics[width=0.45\linewidth]{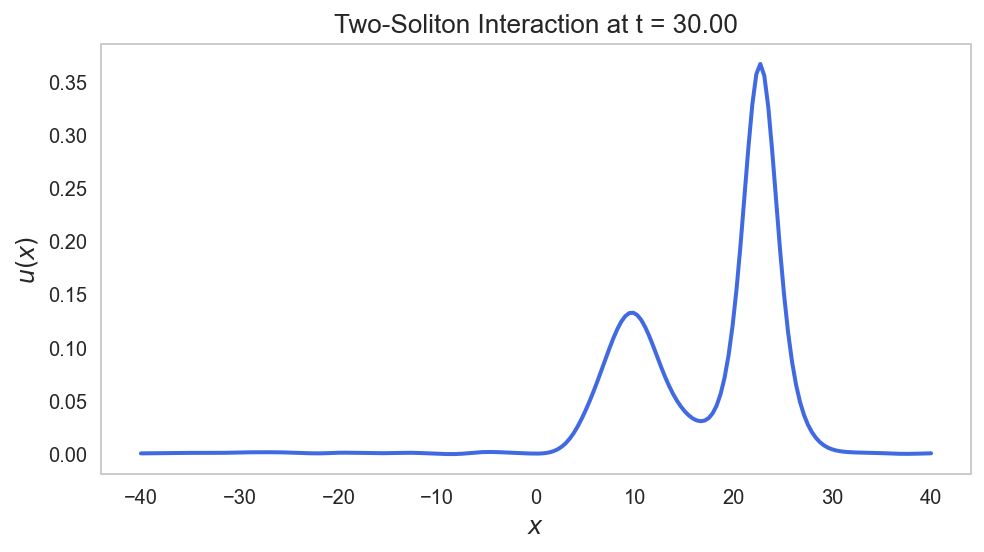}
\end{tabular}
\end{minipage}%
}
\caption{Collision of Two-Soliton Interaction}
\label{fig:eight_box}
\end{figure}
\begin{figure}[H]
    \centering
    \includegraphics[width=1\linewidth]{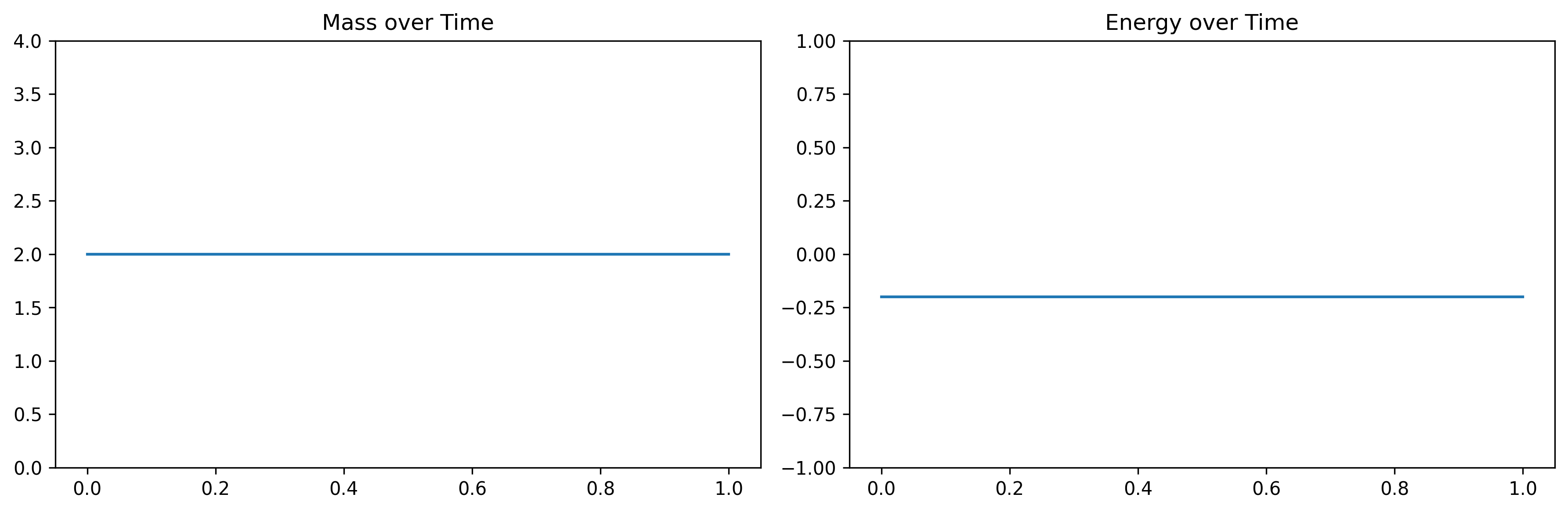}
    \caption{Mass (left) and energy (right) conservation over time for the PINN solution for two soliton interaction.}
    \label{fig:mass_energy2}
\end{figure}
\begin{figure}[H]
    \centering
    \includegraphics[width=1\linewidth]{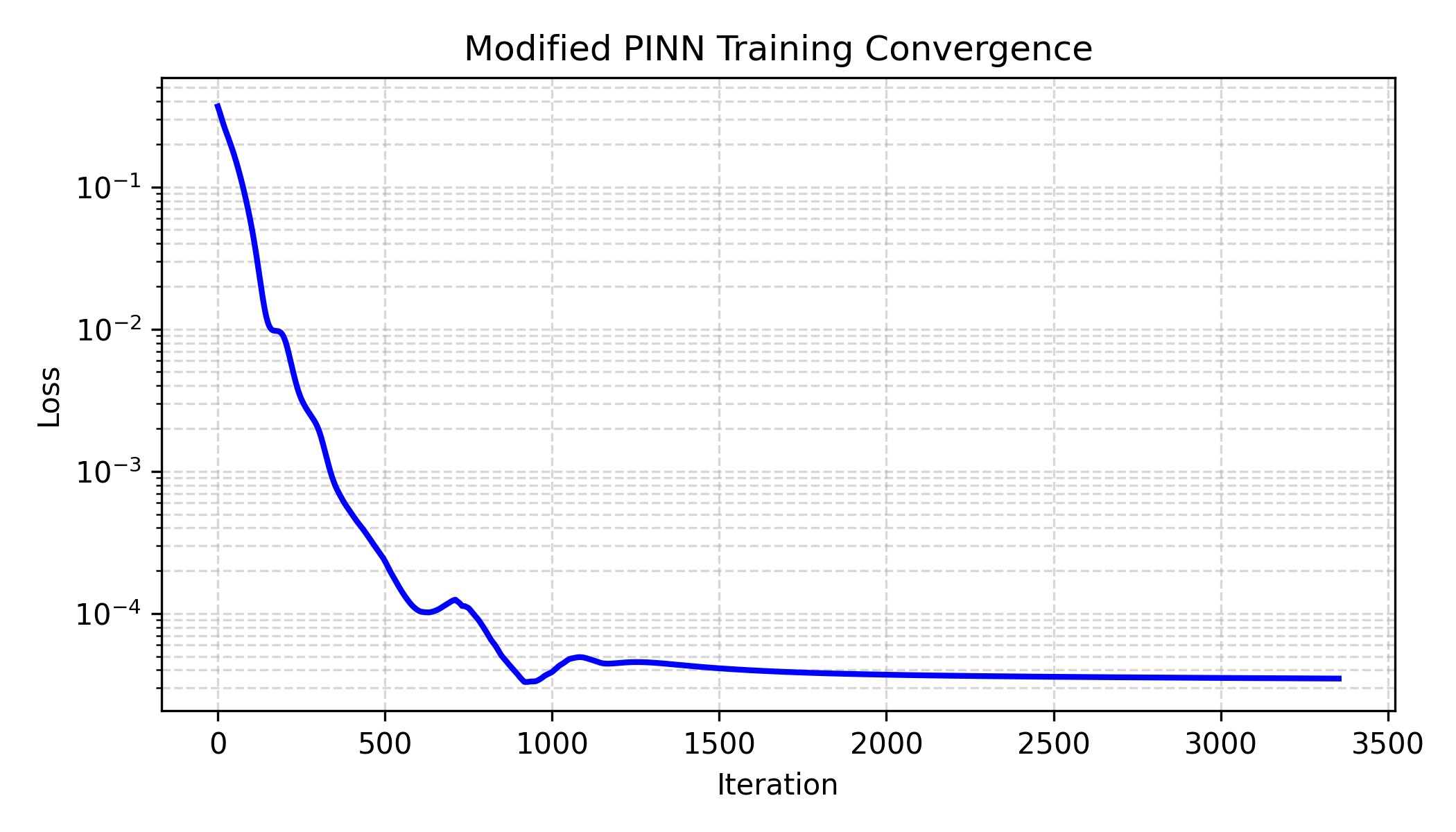}
    \caption{Training profile}
    \label{fig:training1}
\end{figure}
\begin{figure}[H]
    \centering
    \includegraphics[width=1.1\linewidth]{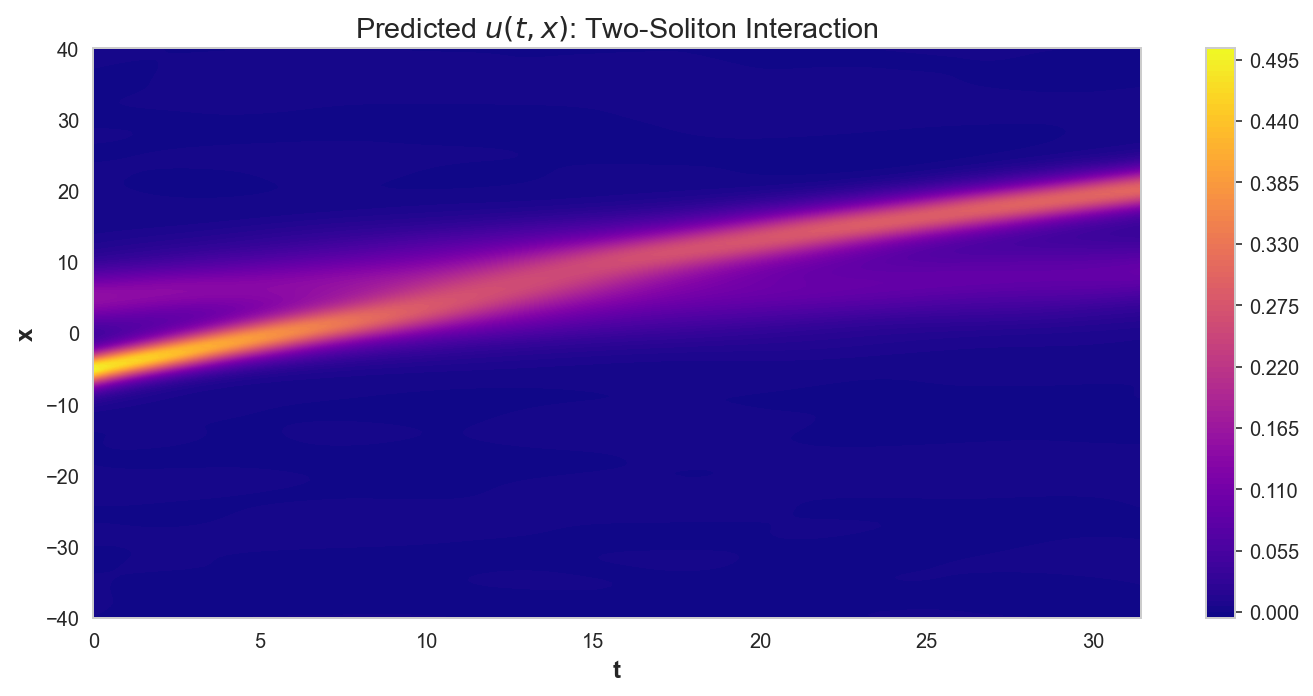 }
    \caption{2D plot of collision of  two soliton}
    \label{fig:surface_plot2}
\end{figure}

\vspace{1ex}
\noindent\textbf{\textit{Two-Soliton Interaction.}} 
Having demonstrated the one-soliton accuracy and convergence behavior in Fig.~\ref{fig:training1}, we now examine whether the proposed method can reproduce strong nonlinear and dispersive phenomena—one of the primary objectives of this study. The two-soliton configuration represents a nonlinear superposition of two solitary waves with distinct amplitudes and velocities governed by the Korteweg–de Vries (KdV) equation. As the faster, larger-amplitude soliton overtakes the slower, smaller one, the waves undergo a brief nonlinear interaction, momentarily merging into an intensified pulse before re-emerging with their original shapes and velocities, except for a characteristic phase shift. This elastic collision, a hallmark of integrable systems, underscores the model’s ability to capture coherent, shape-preserving soliton dynamics. The panels in Figs.~\ref{fig:eight_box} and \ref{fig:surface_plot2} depict this temporal evolution capturing this dynamics, while Fig.~\ref{fig:mass_energy2} confirms that the predicted dynamics preserve the relevant physical invariants

\begin{figure}[H]
\centering
\fbox{%
\begin{minipage}{0.95\linewidth}

    \begin{minipage}[t]{0.48\linewidth}
        \centering
        \includegraphics[width=\linewidth]{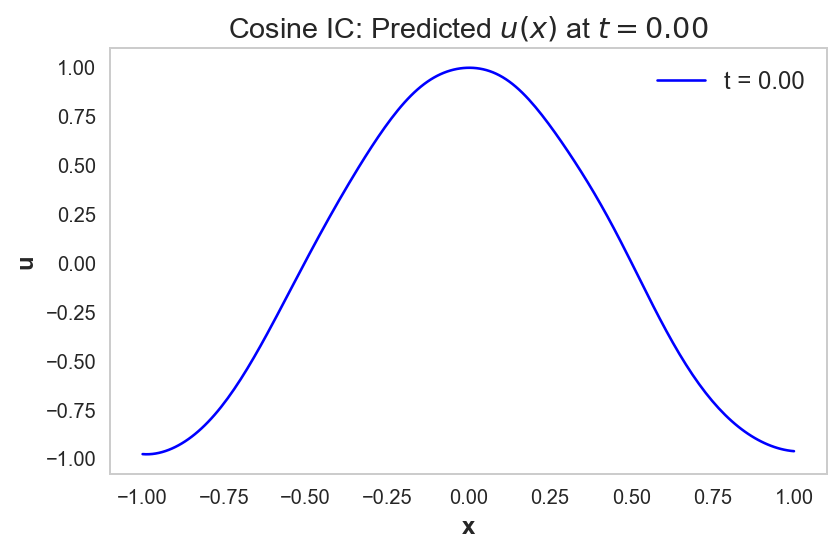}
        \caption*{$t = 0$}
    \end{minipage}
    \hfill
    \begin{minipage}[t]{0.48\linewidth}
        \centering
        \includegraphics[width=\linewidth]{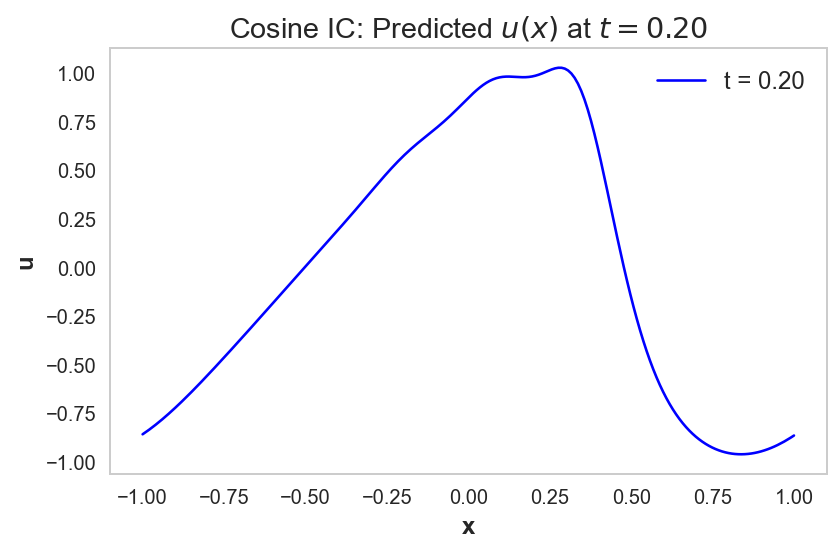}
        \caption*{$t = 0.2$}
    \end{minipage}
    \\[1em]
    \begin{minipage}[t]{0.48\linewidth}
        \centering
        \includegraphics[width=\linewidth]{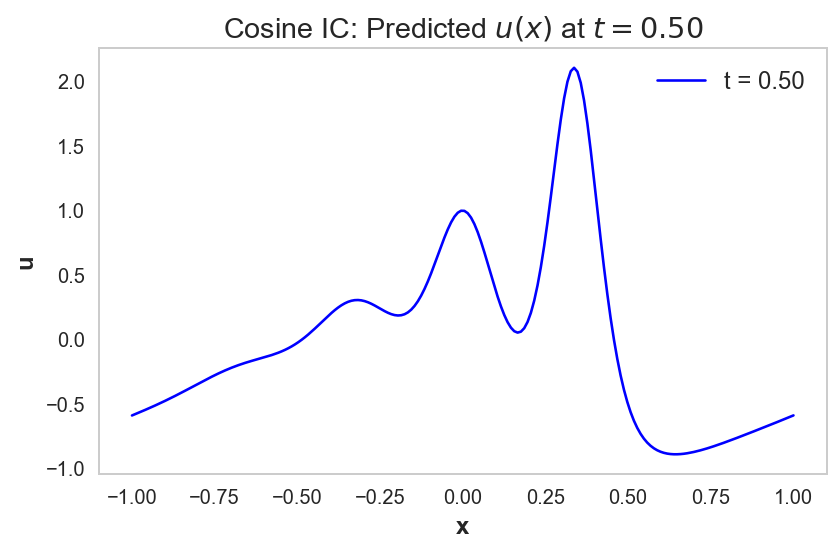}
        \caption*{$t = 0.5$}
    \end{minipage}
    \hfill
    \begin{minipage}[t]{0.48\linewidth}
        \centering
        \includegraphics[width=\linewidth]{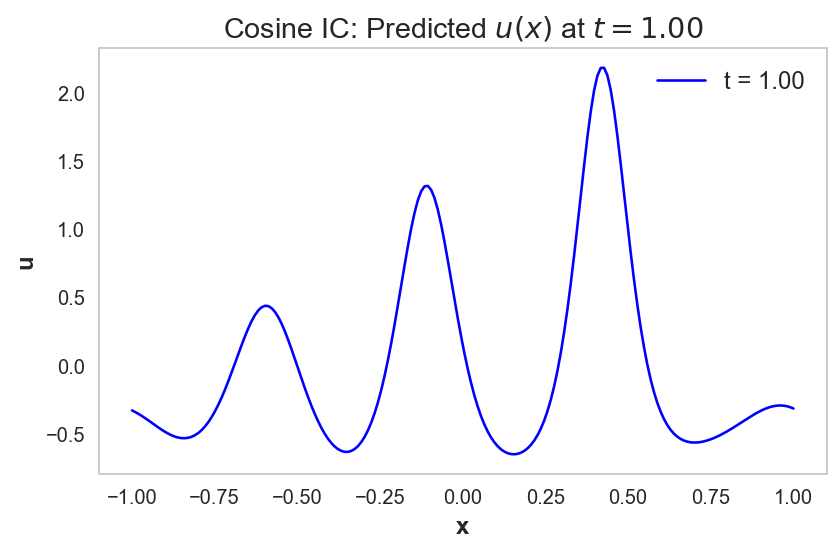}
        \caption*{$t = 1.0$}
    \end{minipage}

\end{minipage}%
}
\caption{Decay of initial pulse into trains of solution}
\label{fig:kdv_decay_trains}
\end{figure}

\begin{figure}[H]
    \centering
    \includegraphics[width=1\linewidth]{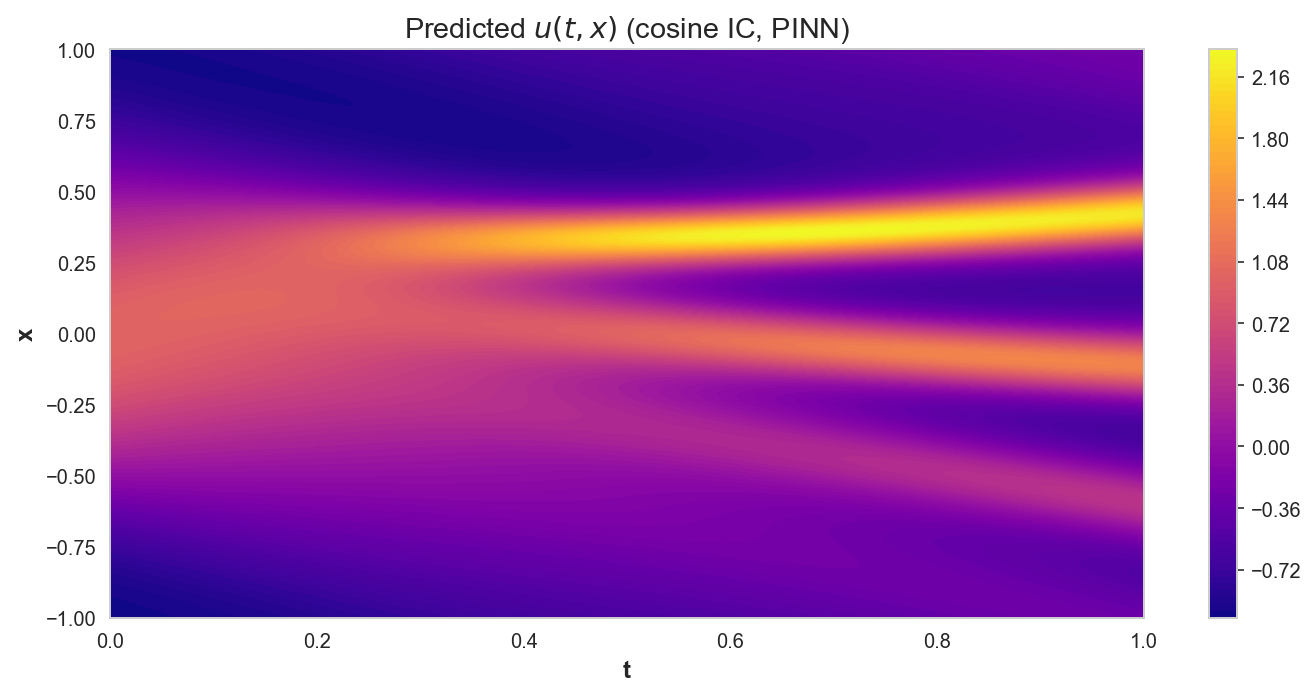}
    \caption{Decay of initial pulse into trains of solution}
    \label{fig:contour3}
\end{figure}

\begin{figure}[H]
    \centering
    \includegraphics[width=1\linewidth]{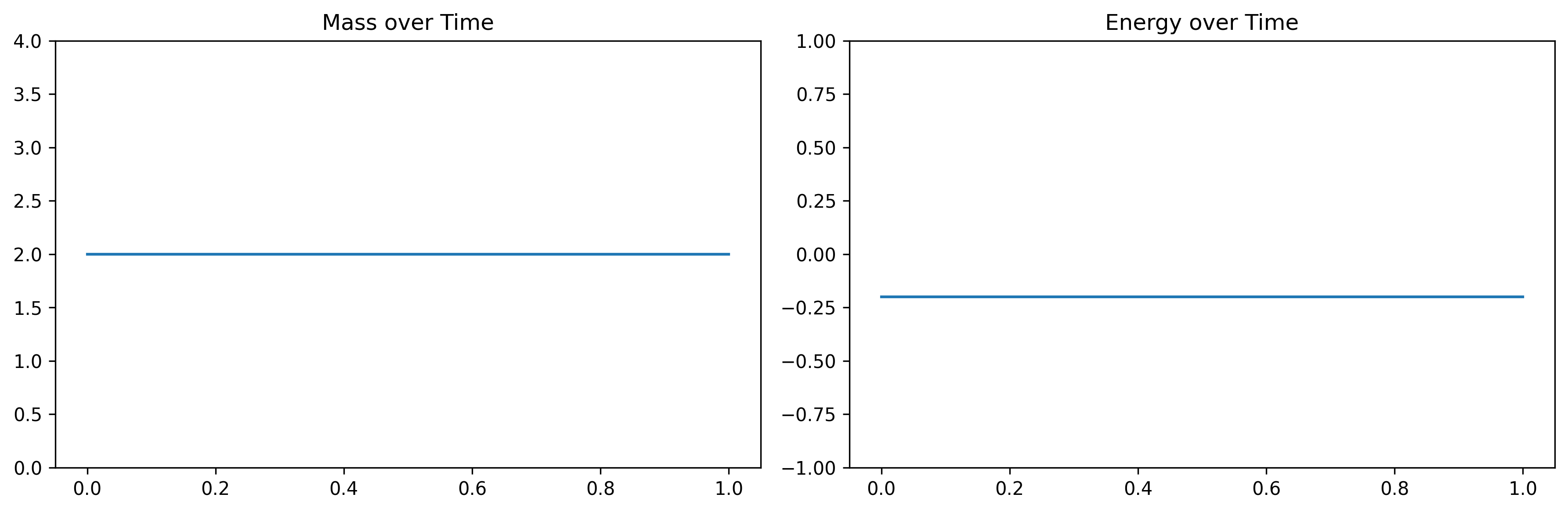}
    \caption{Mass (left) and energy (right) conservation over time for the PINN solution for the cosine profile}
    \label{fig:mass3}
\end{figure}

\noindent\textbf{\textit{Cosine Profile.}} 
To further evaluate the model’s ability to capture complex dispersive dynamics, we initialize the KdV system with a smooth cosine perturbation instead of a soliton profile evolves into a train of solitary-like pulses as the nonlinear and dispersive effects balance over time. As shown in Fig.~\ref{fig:kdv_decay_trains}, ~\ref{fig:contour3} and the temporal snapshots, the trained PINN successfully reproduces this evolution, resolving the formation, steepening, and eventual separation of distinct wave packets. The results demonstrate that the proposed approach can generalize beyond integrable soliton solutions to accurately approximate more general nonlinear wave dynamics governed by the KdV equation while preserving the relevant physical invariants (Fig.~\ref{fig:mass3}).

\begin{figure*}[H]
    \centering
    \includegraphics[width=\textwidth]{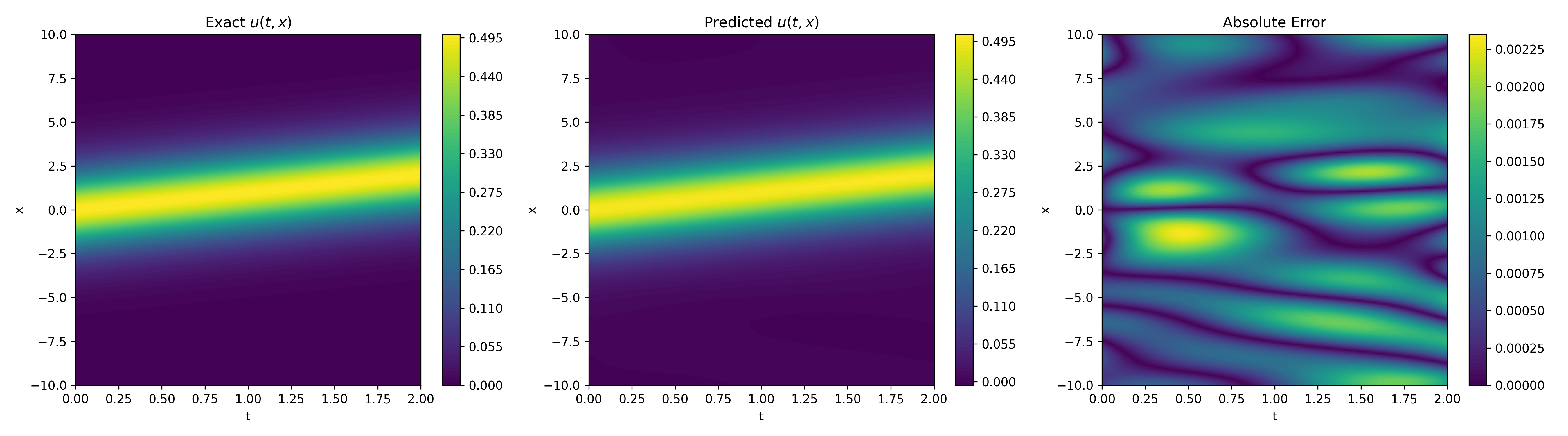}
    \caption{Contour plots of the exact solution $u(t,x)$ (left), PINN-predicted solution (middle), and absolute error (right).}
    \label{fig:contours}
\end{figure*}

\section{Conclusion}
We presented a structure-preserving Physics-Informed Neural Network (SP-PINN) for the Korteweg–de Vries (KdV) equation that embeds the conservation of mass and Hamiltonian energy directly into a single-stage training objective, by introducing \emph{learned invariant weights}—optimized jointly with the network parameters—which act as learnable Lagrange multipliers. Theorem~\ref{thm:exact_conservation} shows that any finite stationary point of this learned-weight objective enforces \emph{exact} conservation of mass and energy, providing a simple theoretical guarantee for invariant preservation. Practically, pairing this loss with sinusoidal activations which isnt explored greatly in literature and L-BFGS captures oscillatory, dispersive dynamics more faithfully than conventional \texttt{tanh}-based PINNs. Across single-soliton propagation, two-soliton interaction, and cosine-pulse breakup, the method reproduces hallmark KdV behaviors while maintaining small invariant errors over long horizons and mitigating the drift observed in vanilla PINNs. Overall, the framework is simple to implement, empirically stable, and aligns the learned solution with the underlying Hamiltonian structure.

\

\noindent\textbf{Limitations.} 
The current experiments are restricted to smooth initial conditions and fixed boundary types; evaluating robustness under rough inputs, stochastic forcing, and noisy measurements is an important next step toward broader applicability. 

\section*{Availability of Data and Materials}
The datasets and code used in this study are available from the corresponding author upon reasonable request.

\section*{Conflict of Interest}
The authors declare that there are no conflicts of interest.

{\small
\bibliographystyle{ieee}
\bibliography{egbib}
}

\section*{Considerations}
\subsubsection*{Learned invariant weights}

We also consider \emph{learned} invariant weights, in which the coefficients of the conservation-law terms are treated as
trainable parameters and optimized jointly with the network.

Specifically, we introduce scalar parameters $\alpha_M, \alpha_E \in
\mathbb{R}$ and define positive weights
\begin{equation}
\lambda_M = e^{\alpha_M}, \qquad \lambda_E = e^{\alpha_E}.
\end{equation}
The total loss then becomes
\begin{equation}
L_{\text{total}}
= L_{\text{IC}} + L_{\text{PDE}} + L_{\text{BC}}
+ \lambda_M L_{\text{mass}} + \lambda_E L_{\text{energy}}.
\label{eq:total_loss_learned}
\end{equation}
The parameters $(\theta, \alpha_M, \alpha_E)$ are optimized jointly via
L-BFGS, with gradients computed by automatic differentiation. The
exponential parameterization guarantees $\lambda_M, \lambda_E > 0$
without imposing hard constraints on $\alpha_M, \alpha_E$.

From a penalty-method perspective, very small values of
$\lambda_M, \lambda_E$ reduce \eqref{eq:total_loss_learned} to a
standard PINN that largely ignores conservation, whereas excessively
large values place almost all emphasis on matching the invariants and
can stall the reduction of the PDE residual. Learning
$\lambda_M$ and $\lambda_E$ allows the optimization to discover an
intermediate regime where both the KdV dynamics and the conservation
laws contribute meaningfully to the objective.

\newtheorem{theorem}{Theorem} \begin{theorem}[Exact conservation at finite stationary points] \label{thm:exact_conservation}
Assume that the loss components $L_{\text{IC}}, L_{\text{PDE}},
L_{\text{BC}}, L_{\text{mass}}, L_{\text{energy}} : \mathbb{R}^p \to
[0,\infty)$ are continuously differentiable with respect to the network
parameters $\theta \in \mathbb{R}^p$. Consider the learned-weight loss
\eqref{eq:total_loss_learned}. Suppose $(\theta^\ast,\alpha_M^\ast,
\alpha_E^\ast)$ is a finite stationary point of $L_{\text{total}}$,
i.e.,
\begin{align*}
  \nabla_\theta L_{\text{total}}(\theta^\ast,\alpha_M^\ast,\alpha_E^\ast) &= 0, \\
  \frac{\partial L_{\text{total}}}{\partial \alpha_M}(\theta^\ast,\alpha_M^\ast,\alpha_E^\ast) &= 0, \\
  \frac{\partial L_{\text{total}}}{\partial \alpha_E}(\theta^\ast,\alpha_M^\ast,\alpha_E^\ast) &= 0,
\end{align*}
with $\alpha_M^\ast,\alpha_E^\ast \in \mathbb{R}$. Then the corresponding
network solution satisfies
\[
  L_{\text{mass}}(\theta^\ast) = 0,
  \qquad
  L_{\text{energy}}(\theta^\ast) = 0.
\]
In particular, any finite stationary point of the learned-weight loss
enforces exact conservation of mass and energy.
\end{theorem}

\begin{proof}
By definition of $L_{\text{total}}$ in \eqref{eq:total_loss_learned}, we
have
\begin{align}
  L_{\text{total}}(\theta,\alpha_M,\alpha_E)
  &= L_{\text{IC}}(\theta) + L_{\text{PDE}}(\theta) + L_{\text{BC}}(\theta) \nonumber \\
  &\quad + e^{\alpha_M} L_{\text{mass}}(\theta)
  + e^{\alpha_E} L_{\text{energy}}(\theta).
\end{align}

Differentiating with respect to $\alpha_M$ and $\alpha_E$ yields
\[
  \frac{\partial L_{\text{total}}}{\partial \alpha_M}
  = e^{\alpha_M} L_{\text{mass}}(\theta), \qquad
  \frac{\partial L_{\text{total}}}{\partial \alpha_E}
  = e^{\alpha_E} L_{\text{energy}}(\theta).
\]
At a finite stationary point $(\theta^\ast,\alpha_M^\ast,\alpha_E^\ast)$
we have
\[
  0 = \frac{\partial L_{\text{total}}}{\partial \alpha_M}
  (\theta^\ast,\alpha_M^\ast,\alpha_E^\ast)
  = e^{\alpha_M^\ast} L_{\text{mass}}(\theta^\ast).
\]
Since $e^{\alpha_M^\ast} > 0$ for all finite $\alpha_M^\ast$, it follows
that $L_{\text{mass}}(\theta^\ast) = 0$. An identical argument applied to
$\partial L_{\text{total}} / \partial \alpha_E$ shows that
$L_{\text{energy}}(\theta^\ast) = 0$. This proves the claim.
\end{proof}

\end{document}